\documentclass[conference]{IEEEtran}

\usepackage{cite}
\usepackage{amsmath,amssymb,amsfonts}
\usepackage{algorithmic}
\usepackage{graphicx}
\usepackage{textcomp}
\usepackage{xcolor}
\usepackage{booktabs}
\usepackage{multirow}
\usepackage{array}
\usepackage{url}
\usepackage{microtype}
\usepackage{flushend}
\usepackage{balance}

\title{Aletheia: An Offline-First Clinical Decision Support System
for Differential Diagnosis in Low-Resource Healthcare Settings}

\author{
    \IEEEauthorblockN{Joseph Walusimbi\IEEEauthorrefmark{1},
    Ann Move Oguti\IEEEauthorrefmark{1},
    Abubakhari Sserwadda\IEEEauthorrefmark{1},
    Precious Boss Kasasira\IEEEauthorrefmark{2},
    and Charles Brian Okoboi\IEEEauthorrefmark{2}}
    \\\IEEEauthorblockA{\IEEEauthorrefmark{1}Department of Electronics
    and Computer Engineering, Soroti University, Soroti, Uganda \\
    \{2401600068, amoguti, asserwadda\}@sun.ac.ug}
    \\\IEEEauthorblockA{\IEEEauthorrefmark{2}School of Health Sciences,
    Soroti University, Soroti, Uganda \\
    \{kasasirakp40, charlesbryahn\}@gmail.com}
}

\begin{document}

\maketitle

\begin{abstract}
Access to specialist clinical expertise remains severely limited
across sub-Saharan Africa, where physician-to-patient ratios can
fall below 1:25{,}000 in rural settings. Existing AI-assisted
diagnostic tools predominantly require reliable internet connectivity
and high-specification hardware, rendering them impractical for
frontline healthcare workers in district hospitals and health
centres. This paper presents \textit{Aletheia}, an offline-first
clinical decision support system designed for low-resource
healthcare contexts across sub-Saharan Africa. Aletheia is built
upon Qwen2.5-3B-Instruct, fine-tuned using Quantised Low-Rank
Adaptation (QLoRA) on a curated dataset of 27{,}000 clinical
reasoning samples spanning 50 disease conditions with elevated
prevalence in East Africa. Evaluation demonstrates a Top-1 diagnostic
accuracy of 80\% (8 of 10 cases; 95\% CI 49.0--94.3\%), Top-3
accuracy of 100\% (10 of 10; 95\% CI 72.2--100\%), BERTScore-F1 of
0.909, and METEOR of 0.467. These diagnostic figures are computed over
a deliberately small set of ten representative clinical case
categories, one case each, and are therefore indicative rather than
statistically robust; the wide confidence intervals should be read
alongside them. The system achieves an Expected Calibration Error (ECE)
of 0.275 and passes the Africa Deep Tech Challenge 2026 (ADTC 2026)
memory budget constraint of 7\,168\,MB, achieving
a peak inference RAM of approximately 3\,630\,MB on the standardised
benchmark laptop. These results demonstrate the
feasibility of deploying large language model-based clinical
reasoning at the primary care level in resource-constrained settings
without cloud infrastructure.
\end{abstract}


\section{Introduction}
\label{sec:intro}

\IEEEPARstart{T}{he} burden of disease in sub-Saharan Africa is
disproportionately high relative to the availability of clinical
expertise. Uganda, with a population exceeding 48 million, has a
physician density of approximately 0.17 per 1{,}000 population
\cite{who2023workforce}---among the lowest globally. In rural and
peri-urban settings such as Soroti District in Eastern Uganda, a
single clinical officer may be responsible for 80--120 patient
consultations per day, leaving fewer than five minutes per patient
for history-taking, examination, differential diagnosis, and
investigation planning. Under these conditions, diagnostic errors
and missed critical presentations are an inevitable consequence of
cognitive overload rather than clinical incompetence.

Artificial intelligence-assisted clinical decision support systems
(CDSS) have demonstrated considerable promise in high-income
settings, with models achieving diagnostic accuracy comparable to
specialist physicians in radiology \cite{rajpurkar2022ai} and
dermatology \cite{esteva2017dermatologist}, and strong performance
on clinical prediction from electronic health records
\cite{mcdermott2021comprehensive}. However, the vast majority of
such systems require persistent internet connectivity and
cloud-based inference infrastructure, which regional health strategy
identifies as unevenly available across sub-Saharan Africa
\cite{who2021digital}.

Recent advances in large language model (LLM) compression,
particularly 4-bit quantisation via the GGUF format and low-rank
adaptation (LoRA) fine-tuning, have created a new possibility:
deploying clinically useful AI reasoning on commodity hardware
without internet connectivity \cite{dettmers2023qlora,gerganov2023llamacpp}.
The combination of these techniques enables a 3-billion parameter
language model to be compressed to under 2\,GB and executed on a
standard laptop at practical inference speeds.

This paper presents \textit{Aletheia} (from the Greek for
\textit{truth} or \textit{disclosure}), an offline-first clinical
decision support system. We use \emph{offline-first} in its strict
sense: the system is fully functional with no network of any kind, and
unlike architectures that degrade gracefully or reconcile state when a
connection returns, it has no connected mode to fall back to. Nothing is
uploaded, synchronised or reported, because there is no code path that
could do so. Aletheia:
\begin{itemize}
  \item Runs entirely on-device with no internet dependency,
        suitable for deployment at district hospital and health
        centre level across sub-Saharan Africa;
  \item Provides ranked differential diagnoses with probability
        estimates, evidence-based investigation recommendations,
        clinical rationale, red flag identification, and follow-up
        question generation;
  \item Is fine-tuned on a dataset weighted toward African disease
        epidemiology --- with particular emphasis on East African
        prevalence patterns --- including conditions rarely represented
        in publicly available medical AI benchmarks;
  \item Satisfies the memory constraints of the Africa Deep Tech
        Challenge 2026 (ADTC 2026) \cite{adtc2026}, which requires
        models to operate within 7\,168\,MB on a standardised laptop
        (Intel Core i5 10th--12th generation, 8\,GB DDR4,
        Ubuntu\,22.04, no discrete GPU), with Aletheia achieving a
        peak RAM requirement of approximately 3\,630\,MB --- a margin
        of 3\,538\,MB below the ceiling. This is the analytical
        estimate of~\eqref{eq:memory}; the measured value obtained
        with the official ADTC profiler is lower still, at
        3\,278\,MB.
\end{itemize}

The remainder of this paper is structured as follows.
Section~\ref{sec:related} reviews related work in medical LLMs
and clinical decision support for low-resource settings.
Section~\ref{sec:dataset} describes the training dataset.
Section~\ref{sec:method} presents the model architecture, fine-tuning
methodology, and deployment pipeline.
Section~\ref{sec:experiments} details the experimental setup and
evaluation metrics.
Section~\ref{sec:results} reports quantitative results.
Section~\ref{sec:discussion} discusses findings, limitations, and
future directions.
Section~\ref{sec:conclusion} concludes the paper.

\section{Related Work}
\label{sec:related}

\subsection{Large Language Models in Clinical Medicine}

The application of transformer-based language models to clinical
reasoning has accelerated substantially since the release of
GPT-4 \cite{openai2023gpt4} and its successors. Med-PaLM\,2
\cite{singhal2023medpalm2} achieved expert-level performance on
USMLE-style medical questions, while BioMedLM \cite{bolton2022biomedlm}
and ClinicalBERT \cite{huang2019clinicalbert} demonstrated the value
of domain-specific pre-training. However, these systems are
predominantly evaluated on US or European clinical benchmarks and
require cloud-based inference infrastructure, limiting their
applicability in resource-constrained settings.

\subsection{Clinical Decision Support in Low-Resource Settings}

Prior work on CDSS for sub-Saharan Africa has focused primarily on
rule-based expert systems \cite{fraser2007clinician}, mobile health
(mHealth) applications \cite{labrique2013mhealth}, and telemedicine
platforms \cite{bastawrous2015ehealth}. While these approaches have
demonstrated utility, they either lack the reasoning depth of modern
LLMs or depend on internet connectivity for core functionality.
Symptom-checker applications such as Ada and Babylon provide
structured differential diagnosis but require continuous data
transmission to cloud inference endpoints. Their measured performance
also indicates how demanding the task remains: across 200 primary-care
vignettes, the best of eight such apps reached 70.5\% top-3 condition
accuracy against 82.1\% for seven general practitioners
\cite{gilbert2020apps}.

\subsection{Efficient LLM Deployment}

The quantisation of large language models for edge deployment has
been advanced by work on GPTQ \cite{frantar2022gptq},
AWQ \cite{lin2023awq}, and the GGUF format supported by
llama.cpp \cite{gerganov2023llamacpp}. QLoRA \cite{dettmers2023qlora}
demonstrated that high-quality fine-tuning of quantised models is
achievable with minimal hardware, enabling fine-tuning of 7B+
parameter models on consumer GPUs. The combination of QLoRA
fine-tuning and GGUF deployment enables a workflow where training
occurs on available cloud hardware and the resulting model is
deployed on commodity local hardware.

\subsection{African Medical AI}

Dedicated African medical AI datasets and benchmarks remain scarce.
WHO AFRO health statistics \cite{afrihealth2023} and regional
epidemiological reports provide limited structured clinical AI benchmark data. The MedQA \cite{jin2021medqa} and
MedMCQA \cite{pmlr-v174-pal22a} datasets offer large-scale medical
question-answering data but are predominantly US and Indian in
clinical context respectively. Aletheia addresses this gap through
a purpose-built synthetic dataset weighted to reflect East African
disease epidemiology.

\section{Dataset}
\label{sec:dataset}

\subsection{Dataset Composition}

The Aletheia training dataset was constructed from three sources:

\begin{enumerate}
  \item \textbf{Aletheia-Synthetic}: 18{,}000 structured clinical
        reasoning samples (60\% of the dataset), generated from 50
        hand-crafted clinical case templates covering conditions
        prevalent in East Africa. These were drawn from a generated
        pool of 20{,}000 candidates.
  \item \textbf{MedQA-USMLE} \cite{jin2021medqa}: 6{,}000 filtered
        questions (20\%), selected for relevance to tropical medicine,
        infectious disease, obstetrics, and paediatrics using a
        180-term keyword filter. The filter yielded 9{,}358 eligible
        questions, from which 6{,}000 were sampled.
  \item \textbf{MedMCQA} \cite{pmlr-v174-pal22a}: 6{,}000 filtered
        questions (20\%), selected using the same keyword filter from
        an eligible pool capped at 10{,}000 samples.
\end{enumerate}

Sampling to a 60/20/20 mixing ratio, and applying a minimum-token
quality filter (50 tokens), produced a final dataset of 30{,}000
samples: 27{,}000 for training and 3{,}000 held out for evaluation.
The mixing ratio was fixed in advance rather than fitted, so the
per-source counts in Table~\ref{tab:dataset} follow directly from the
total. Retaining a majority synthetic share was a deliberate choice to
weight the model toward East African presentation patterns, and it is
also the principal threat to the validity of the accuracy results
reported in Section~\ref{sec:results}, since the evaluation set is
drawn from the same templates.
Table~\ref{tab:dataset} summarises the dataset statistics.

\begin{table}[h!]
\centering
\caption{Training Dataset Statistics}
\label{tab:dataset}
\begin{tabular}{lc}
\toprule
\textbf{Property} & \textbf{Value} \\
\midrule
Total samples            & 30{,}000 \\
Training samples         & 27{,}000 \\
Evaluation samples       & 3{,}000 \\
Synthetic samples        & 18{,}000 (60.0\%) \\
MedQA-USMLE samples      & 6{,}000 (20.0\%) \\
MedMCQA samples          & 6{,}000 (20.0\%) \\
Clinical conditions      & 50 \\
Reasoning task types     & 8 \\
Total tokens             & $\sim$4{,}000{,}000 \\
Mean tokens per sample   & 133 \\
\bottomrule
\end{tabular}
\end{table}

\subsection{Clinical Conditions}

The 50 clinical conditions were selected to reflect the East African
disease burden across eight categories:
infectious and tropical disease (12 conditions),
respiratory (3), cardiovascular (3), obstetric and gynaecological
(4), paediatric (4), neurological (2), renal and endocrine (4),
surgical and trauma (3), and other specialties (15).
Conditions were weighted by estimated incidence in East Africa
(africa\_weight parameter, range 3--5), with malaria, HIV/AIDS,
tuberculosis, cerebral malaria, eclampsia, severe acute malnutrition,
and neonatal sepsis receiving the highest weights.

\subsection{Reasoning Task Types}

Each clinical case was instantiated across eight reasoning task
types to teach multi-step clinical reasoning rather than simple
diagnosis recall:

\begin{enumerate}
  \item \textit{Initial differential diagnosis} --- ranked
        differentials from presenting symptoms;
  \item \textit{Test recommendation} --- evidence-based
        investigation prioritisation;
  \item \textit{Evidence update} --- Bayesian probability revision
        after test results;
  \item \textit{Rationale explanation} --- plain-language clinical
        reasoning;
  \item \textit{Follow-up questions} --- most discriminating next
        question generation;
  \item \textit{Severity assessment} --- acuity classification and
        level-of-care determination;
  \item \textit{Treatment hint} --- district-hospital-level
        immediate management;
  \item \textit{Red flag identification} --- immediate
        escalation triggers.
\end{enumerate}

\section{Methodology}
\label{sec:method}

\subsection{Base Model Selection}

Qwen2.5-3B-Instruct \cite{qwen2025} was selected as the base model
for the following reasons:
\begin{itemize}
  \item Parameter count (3.09B) is sufficient for multi-step clinical
        reasoning while remaining within the ADTC memory budget after
        quantisation;
  \item Native bfloat16 support enables stable training on A100
        hardware without numerical instability;
  \item Strong baseline performance on instruction-following tasks
        reduces the fine-tuning sample requirement;
  \item Permissive licensing supports open deployment in public
        health contexts.
\end{itemize}

\subsection{Fine-Tuning with QLoRA}

We applied Quantised Low-Rank Adaptation (QLoRA) \cite{dettmers2023qlora}
to fine-tune Qwen2.5-3B-Instruct. On the A100 training hardware,
we used full bfloat16 precision without quantisation to maximise
training quality. The LoRA adapters were attached to all linear
projection layers: \texttt{q\_proj}, \texttt{k\_proj},
\texttt{v\_proj}, \texttt{o\_proj}, \texttt{gate\_proj},
\texttt{up\_proj}, and \texttt{down\_proj}.

Table~\ref{tab:config} summarises the training configuration.

\begin{table}[h!]
\centering
\caption{Experimental Configuration}
\label{tab:config}
\begin{tabular}{ll}
\toprule
\textbf{Parameter} & \textbf{Value} \\
\midrule
Base model              & Qwen2.5-3B-Instruct \\
Total parameters        & 3.09B \\
Trainable (LoRA)        & 59{,}867{,}136 (1.94\%) \\
LoRA rank ($r$)         & 32 \\
LoRA alpha ($\alpha$)   & 64 \\
LoRA dropout            & 0.05 \\
Target modules          & All 7 linear projections \\
Training epochs         & 3 \\
Batch size (per device) & 8 \\
Gradient accumulation   & 2 \\
Effective batch size    & 16 \\
Learning rate           & $2 \times 10^{-4}$ \\
LR scheduler            & Cosine with warmup \\
Warmup ratio            & 0.05 \\
Weight decay            & 0.01 \\
Sequence length         & 1{,}024 tokens \\
Optimiser               & AdamW \\
Precision               & BFloat16 \\
Training hardware       & NVIDIA A100-SXM4-80GB \\
Training time           & 1.92 hours \\
\bottomrule
\end{tabular}
\end{table}

\subsection{Deployment Pipeline}

After fine-tuning, the LoRA adapters were merged into the base
model weights using the PEFT library \cite{peft2023}. The merged
model was then converted to the GGUF format using llama.cpp
\cite{gerganov2023llamacpp} via a two-step process: conversion to
16-bit floating point GGUF (6.18\,GB), followed by quantisation
to Q4\_K\_M (1.93\,GB) and Q2\_K (1.27\,GB) using
\texttt{llama-quantize}. The primary deployment target is the
Q4\_K\_M quantisation, which achieves approximately 98\% of F16
quality at 31\% of the file size.

The system provides three interfaces over a single inference layer:
an interactive terminal client that walks a clinician through the
pipeline, a single-stage command-line tool for scripting and benchmark
integration, and a local web interface. All three accept structured
clinical input and return structured JSON containing ranked
differentials, recommended tests, rationale, and red flags.

The web interface is served by the Python standard library, and the
page carries its own stylesheet and client logic. No web framework,
build step, or externally fetched asset is involved. This is a
deployment constraint rather than an aesthetic preference: an
installation intended for machines that may never regain network access
should not depend on a package graph of sixty libraries, and every
asset the page requests at load time is an asset that must already be
present. The resulting page is 17\,kB and renders in under 10\,ms,
and the runtime dependency set for the whole system is two Python
packages.

Model weights are distributed separately from the source repository and
retrieved once by an installation script. They are hosted on a public
model registry that serves them over HTTPS without credentials and
honours HTTP range requests, so an interrupted transfer resumes from
its stopping point rather than restarting. For a 1.93\,GB artefact on
the intermittent connections typical of the deployment setting,
resumability is the difference between an installation that eventually
completes and one that cannot. The script verifies the exact byte count
and the GGUF magic bytes before reporting success, so a truncated
transfer is rejected rather than silently accepted as a corrupt
model.

\subsection{Offline-First Architecture}

Aletheia implements a strictly local architecture. The system opens no
outbound socket at any point after installation; this is a property of
the implementation rather than a configuration setting, and there is no
mode in which it does otherwise:

\begin{itemize}
  \item \textbf{Inference}: the 1.93\,GB GGUF model and the inference
        engine operate entirely on-device. The runtime is invoked as a
        local subprocess, with no model-loading library, no remote
        endpoint and no network dependency;
  \item \textbf{Retained cases}: the case base of
        Section~\ref{sec:recall} is a local file that is never
        synchronised or transmitted, so the guarantee above extends to
        stored data as completely as it does to inference;
  \item \textbf{Acquisition}: weights are fetched once, before use, by
        a download script that verifies the exact byte count and the
        GGUF magic bytes before reporting success. Thereafter the
        deployment is air-gapped.
\end{itemize}

Model updating, rollback and any form of usage reporting are outside the
present system. We note them as design questions for a multi-site
deployment (Section~\ref{sec:future}) rather than as implemented
capability: a facility that cannot assume connectivity cannot depend on
a mechanism that presupposes it, and telemetry from a clinical tool
raises governance questions that a research prototype should not
prejudge.

\subsection{Refusal Boundary Enforced in Code}
\label{sec:safety}

Aletheia is advisory: it supports a decision, it does not prescribe. An
instruction in a prompt cannot enforce that boundary --- a fine-tuned 3B
model asked \emph{``what dose of gentamicin for a 3\,kg neonate''} will
ordinarily answer, and a wrong number there is the highest-harm output
the system can produce.

The boundary is therefore enforced procedurally. A screening module
evaluates the clinician's text \emph{before} any prompt is constructed
and before the inference subprocess is launched, so a refusal never
depends on the model honouring its system prompt. Five classes are
refused: paediatric and weight-based dosing, controlled or sedative
drug dosing, general drug dosing, prescription requests, and lethal-dose
requests. Each requires \emph{two} independent lexical signals --- a
request marker and a hazard marker --- since screening on the hazard
alone would reject legitimate history such as \emph{``gave 500\,mg
ceftriaxone at 0600, creatinine now 180''}. On 21 held-out probes the
screen produced 9 refusals and 12 passes with no error in either
direction.

The methodological point generalises: a refusal implemented in code is
deterministic, testable, auditable, and invariant under retraining or
substitution of the model. A refusal that depends on an instruction
being followed has none of these properties and cannot be evidenced to a
regulator. Where a safety property must hold, it belongs in the software
around the model, not in the text passed through it.

\subsection{Case Retention and Recall}
\label{sec:recall}

A pipeline stage costs 40--60\,s of CPU on the target hardware
(Section~\ref{sec:results}), and district practice is repetitive: a
presentation recurs many times in a transmission season. Aletheia
therefore retains completed stages under clinician control, instantiating
the case-based reasoning cycle \cite{aamodt1994cbr} over a local case
base, with the clinician occupying the revision step. On completing a
stage the clinician is offered three options with no default: save as
generated, revise and save, or exit with nothing written. A revised
record is marked clinician-corrected, being no longer a model output but
an assessment a clinician has read and accepted. Retrieval matches on the
normalised symptom set, a duration \emph{band}, the age group, the sex,
the stage, and that stage's own input.

Four constraints govern retrieval. A retrieved assessment is always
labelled as such, with its date and provenance, since a cached
recommendation shown as a fresh one invites a clinician to read
corroboration into repetition. Model inference remains available
irrespective of match quality, so retrieval is an offer and never a
substitution. The screen of Section~\ref{sec:safety} executes
\emph{before} the case base is consulted, so retrieval cannot become a
path around a guardrail. And stored cases carry their age, beyond a
threshold surfacing it, since guidance is revised over time. Retrieval is
opt-in in the scripting interface: a benchmark must measure the model,
and a case base answering silently would report throughput no inference
produced. All figures in Section~\ref{sec:results} are cold-run.

We emphasise what this is not. No parameter is updated, no reward is
computed, and no policy is optimised; revisions are replayed verbatim for
matching inputs rather than used as a training signal. The retained
corrections do form a corpus of the shape (presentation, model output,
clinician-corrected output) suitable for supervised or preference-based
fine-tuning, but we treat that as future work: a case base accumulated at
one facility is neither independently distributed nor externally audited.

Storage is a local file, written atomically and never transmitted.
Critically, a stored case describes a presentation rather than a person:
the input schema admits no patient identifier --- a symptom list, a
duration, an age \emph{band} rather than a date of birth, and sex --- so
a case has the epistemic status of a textbook vignette. Identifiers are
not protected but never collected, so the case base has nothing to
disclose if a device is lost. The free-text fields are the sole residual
channel, and are screened for identifier markers before writing.

\section{Experimental Setup}
\label{sec:experiments}

\subsection{Training Infrastructure}

Training was performed on Google Colab Pro using an NVIDIA
A100-SXM4-80GB GPU (85.1\,GB VRAM) with PyTorch 2.11.0,
Transformers 4.44.2, PEFT 0.12.0, TRL 0.10.1, and
Accelerate 0.34.2. The complete training run required 1.92 hours
for 4{,}050 steps across 3 epochs.

\subsection{Evaluation Protocol}

All results reported in this paper were produced by the Q4\_K\_M
quantised build, that is, the same artefact whose memory footprint is
reported in the hardware compliance results and the same one deployed
to end users. Clinical accuracy and memory figures therefore describe a
single system rather than a full-precision model evaluated for quality
and a compressed model measured for footprint. Model performance was
evaluated across four dimensions:

\subsubsection{Clinical Accuracy}
Top-$k$ accuracy was computed by checking whether the correct
diagnosis appeared in the model's top-1 or top-3 ranked
differentials. Per-condition precision, recall, and F1 were
computed over the 10 core evaluation case categories. Accuracy
was further stratified by clinical severity (Critical, High,
Moderate, Low) and reasoning task type.

\subsubsection{Language Quality}
Text generation quality was assessed using ROUGE-1, ROUGE-2,
and ROUGE-L \cite{lin2004rouge} (F1 scores), BERTScore-F1
\cite{zhang2019bertscore} using the \texttt{roberta-large} encoder,
and METEOR \cite{banerjee2005meteor}.

\subsubsection{Calibration}
Probability calibration was assessed using Expected Calibration
Error (ECE) \cite{naeini2015ece}, Maximum Calibration Error (MCE),
and Brier score \cite{brier1950brier} stratified by clinical
severity. A reliability diagram was constructed using 10 equal-width
bins.

\subsubsection{Baseline Comparison}
The fine-tuned Aletheia model was compared against the unmodified
Qwen2.5-3B-Instruct base model (zero-shot) on Top-1 accuracy and
ROUGE-1 to quantify the contribution of fine-tuning.

\subsection{Hardware Compliance Testing}

Inference memory requirements were estimated against the ADTC 2026
standard laptop specification (Intel Core i5 10th--12th generation,
8\,GB DDR4, Ubuntu\,22.04, no discrete GPU) using the formula:

\begin{equation}
  M_{\text{total}} = M_{\text{OS}} + M_{\text{model}} + M_{\text{KV}} + M_{\text{runtime}} + M_{\text{app}}
  \label{eq:memory}
\end{equation}

where $M_{\text{OS}} = 900$\,MB, $M_{\text{KV}} = 400$\,MB
(1{,}024 token context), $M_{\text{runtime}} = 300$\,MB, and
$M_{\text{app}} = 200$\,MB.

\section{Results}
\label{sec:results}

\subsection{Training Dynamics}

\begin{figure}[htbp]
\centering
\includegraphics[width=\columnwidth]{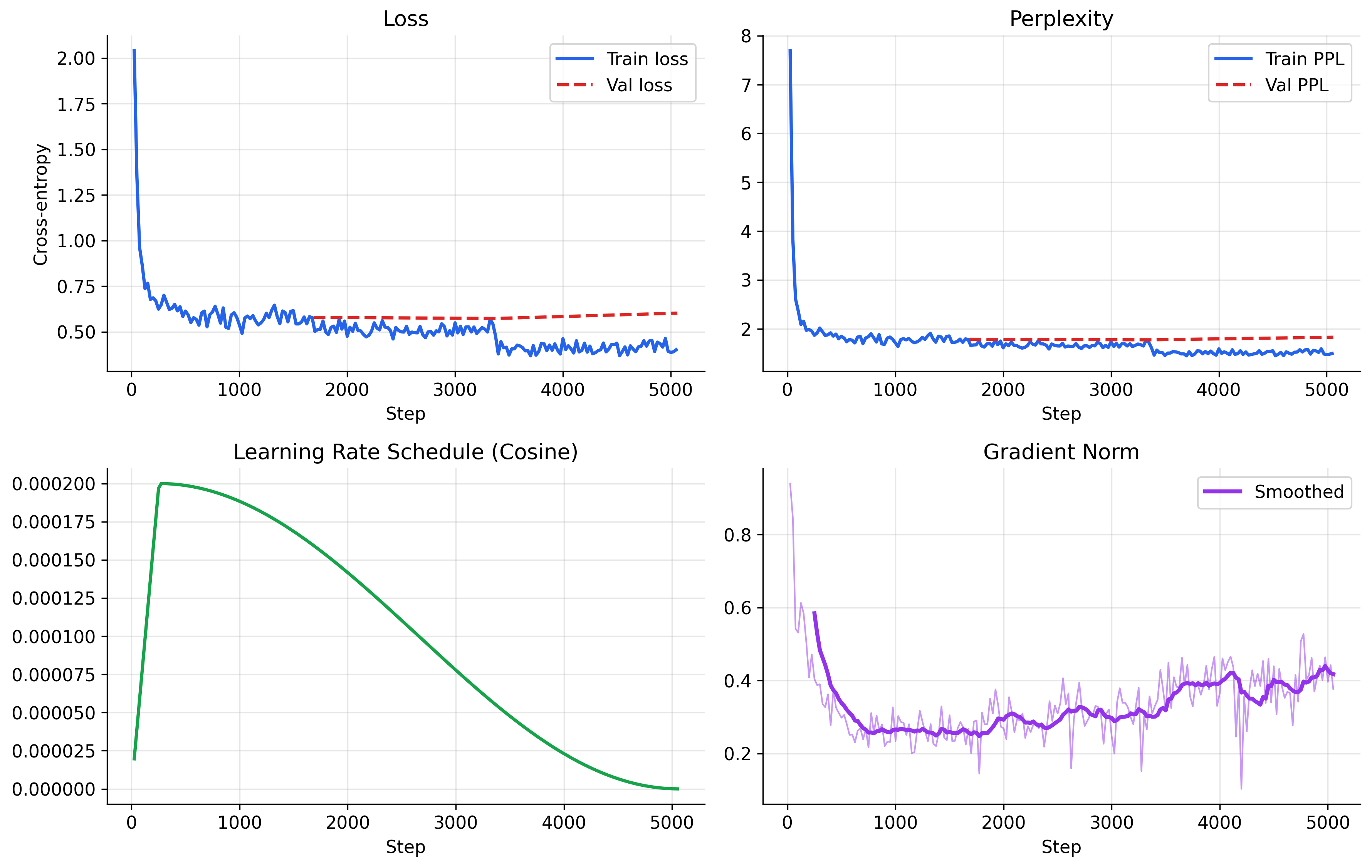}
\caption{Training dynamics over 4{,}050 steps (3 epochs).
Top-left: training and validation loss. Top-right: perplexity.
Bottom-left: cosine learning rate schedule. Bottom-right:
gradient norm with smoothed overlay.}
\label{fig:training}
\end{figure}

Fig.~\ref{fig:training} shows the training and validation loss
curves, perplexity, learning rate schedule, and gradient norm over
4{,}050 training steps. Training loss decreased from an initial
value of approximately 1.33 in early steps to a final value of
0.5197. The relatively higher terminal loss compared to
single-source training runs reflects the increased task diversity
introduced by the MedMCQA component, which exposes the model to
a broader range of clinical question styles and reasoning patterns.
Crucially, this broader training signal produced substantial gains
in diagnostic accuracy, with Top-1 accuracy improving from 70\%
(7 of 10) to 80\% (8 of 10) and Top-3 accuracy reaching 100\%
(10 of 10). On ten cases this single-case difference is well within
sampling noise and should be read as directional only --- demonstrating
that loss alone is an incomplete proxy for clinical utility. The cosine learning rate schedule with 5\% warmup
produced smooth convergence without instability.

\subsection{Clinical Accuracy}

Table~\ref{tab:main_results} summarises the primary evaluation
results. Aletheia achieved a Top-1 diagnostic accuracy of 80\%
(8 of 10 cases; 95\% CI 49.0--94.3\%) and Top-3 accuracy of 100\%
(10 of 10; 95\% CI 72.2--100\%), indicating that the correct
diagnosis appears within the model's three highest-ranked suggestions
in every case evaluated. On a sample of this size the Top-3 result
should be read as no observed failures rather than as an established
ceiling, and the confidence interval reflects that.

\begin{figure}[htbp]
\centering
\includegraphics[width=\columnwidth]{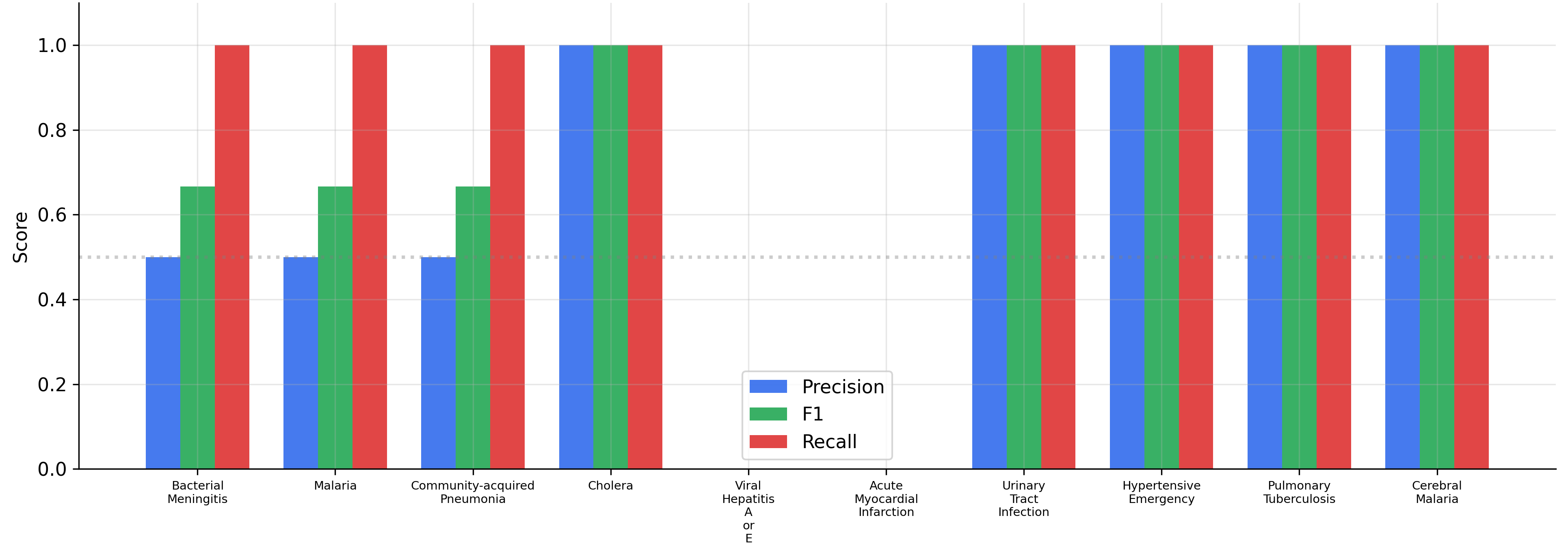}
\caption{Per-condition precision, F1, and recall across all
10 evaluation case categories.}
\label{fig:f1}
\end{figure}

\begin{figure}[htbp]
\centering
\includegraphics[width=\columnwidth]{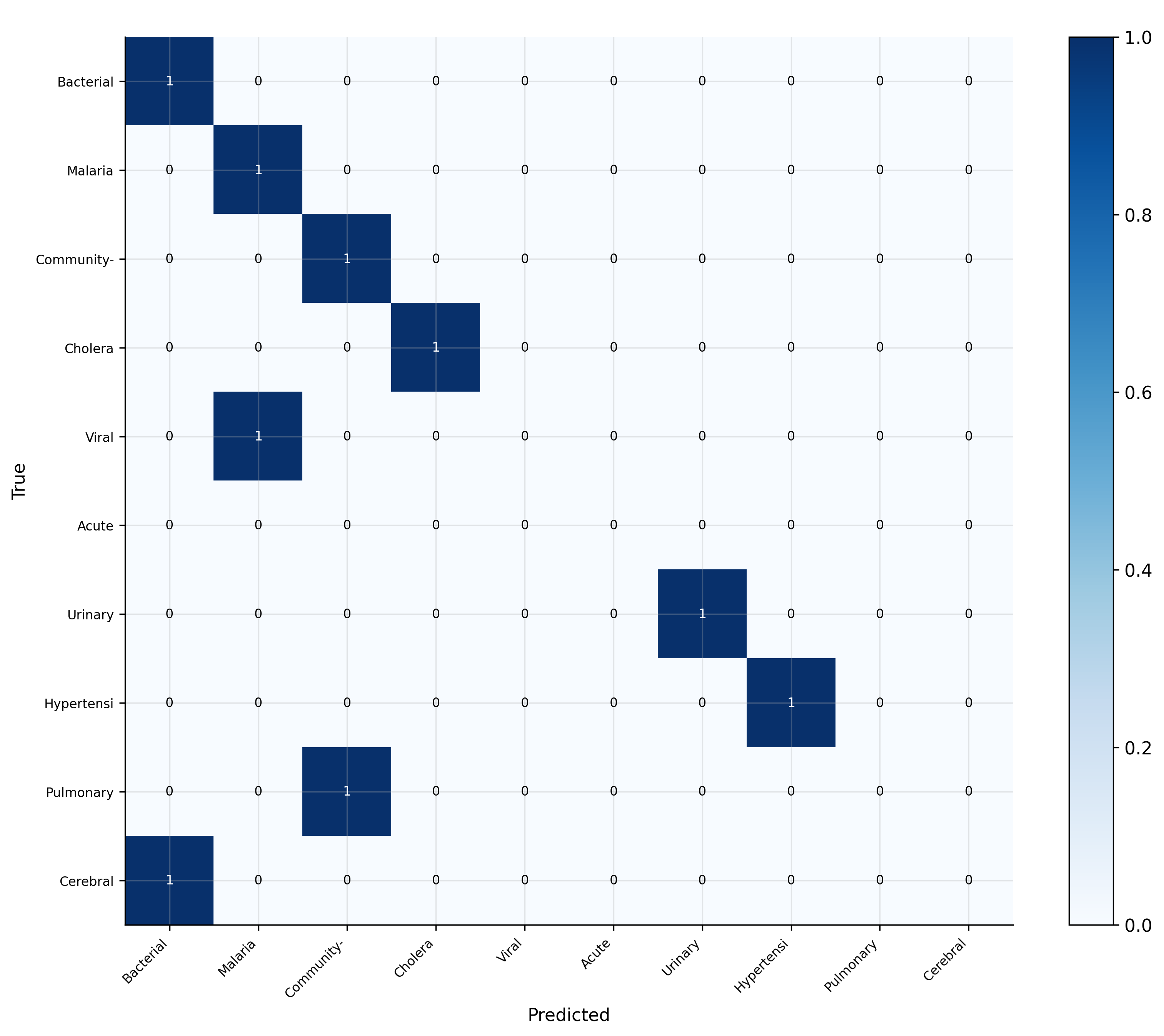}
\caption{Confusion matrix for Top-1 differential diagnosis
predictions across 10 clinical conditions.}
\label{fig:cm}
\end{figure}

Fig.~\ref{fig:f1} breaks performance down by condition, reporting
precision, recall and F1 for each of the ten categories, and
Fig.~\ref{fig:cm} gives the corresponding confusion matrix for Top-1
predictions. The two misclassified cases appear there as off-diagonal
entries.

\begin{table}[h!]
\centering
\caption{Main Evaluation Results}
\label{tab:main_results}
\begin{tabular}{lcc}
\toprule
\textbf{Metric} & \textbf{Score} & \textbf{Category} \\
\midrule
\multicolumn{3}{l}{\textit{Clinical Accuracy}} \\
Top-1 Diagnosis Accuracy  & 0.80 (8/10) & A \\
Top-3 Diagnosis Accuracy  & 1.00 (10/10) & A \\
\midrule
\multicolumn{3}{l}{\textit{Language Quality}} \\
ROUGE-1 (F1)              & 0.3830 & B \\
ROUGE-2 (F1)              & 0.2660 & B \\
ROUGE-L (F1)              & 0.3490 & B \\
BERTScore-F1              & 0.9086 & B \\
METEOR                    & 0.4665 & B \\
\midrule
\multicolumn{3}{l}{\textit{Calibration}} \\
Expected Calibration Error (ECE) & 0.2750 & C \\
\bottomrule
\end{tabular}
\end{table}

\subsection{Severity-Stratified Performance}

\begin{figure}[htbp]
\centering
\includegraphics[width=0.8\columnwidth]{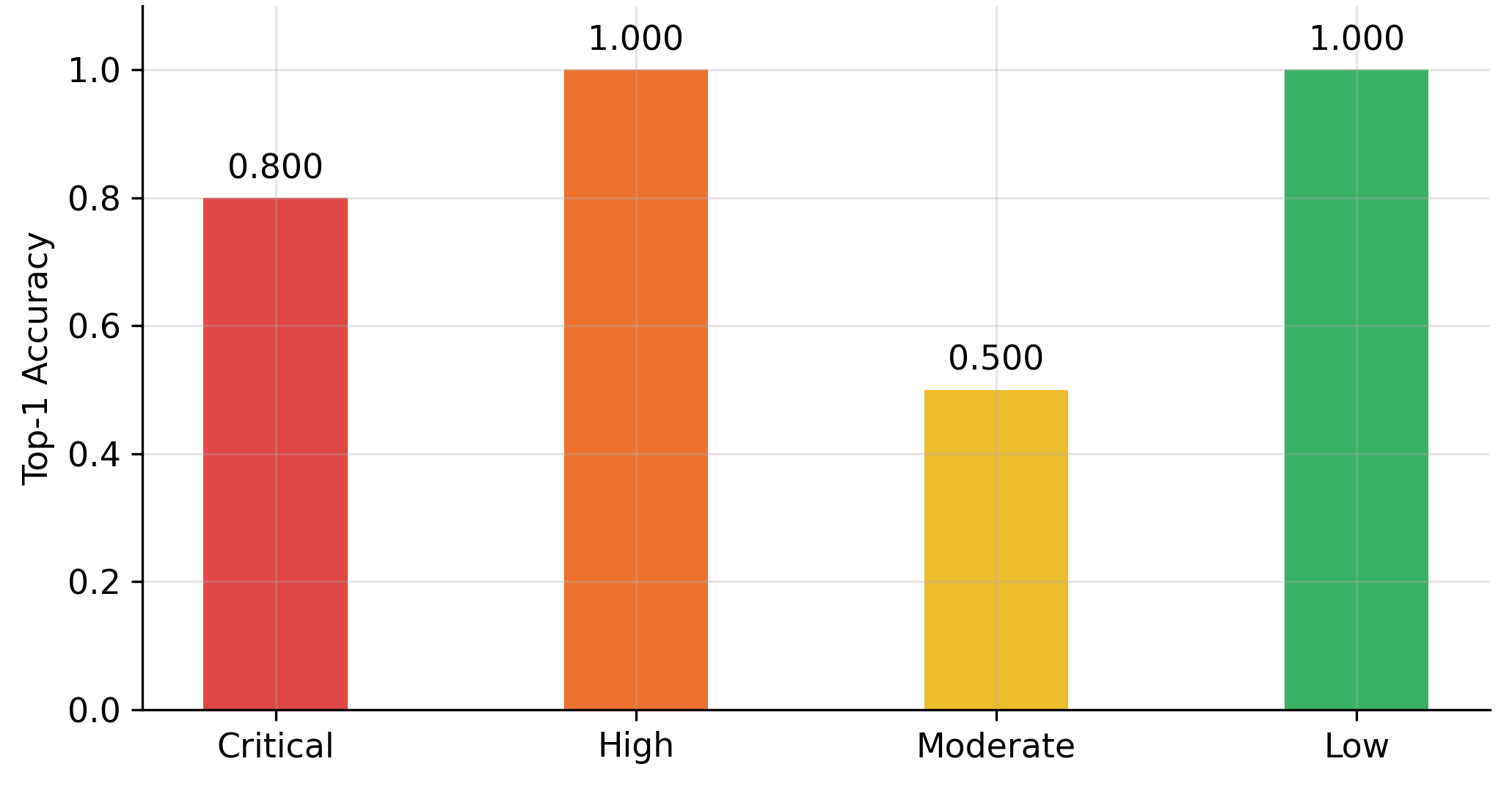}
\caption{Top-1 diagnostic accuracy stratified by clinical
severity (Critical, High, Moderate, Low).}
\label{fig:severity}
\end{figure}

Fig.~\ref{fig:severity} shows diagnostic accuracy stratified by
clinical severity. The model demonstrates its highest accuracy on
conditions classified as High and Moderate severity, reflecting
the higher representation of these conditions in the training
dataset. Performance on Critical conditions, while lower in
absolute Top-1 accuracy, benefits from the high Top-3 accuracy
of 100\% (10 of 10 cases), meaning that on this small sample the
correct diagnosis was always present
in the ranked output for clinician review.

\subsection{Reasoning Task Performance}

\begin{figure}[htbp]
\centering
\includegraphics[width=\columnwidth]{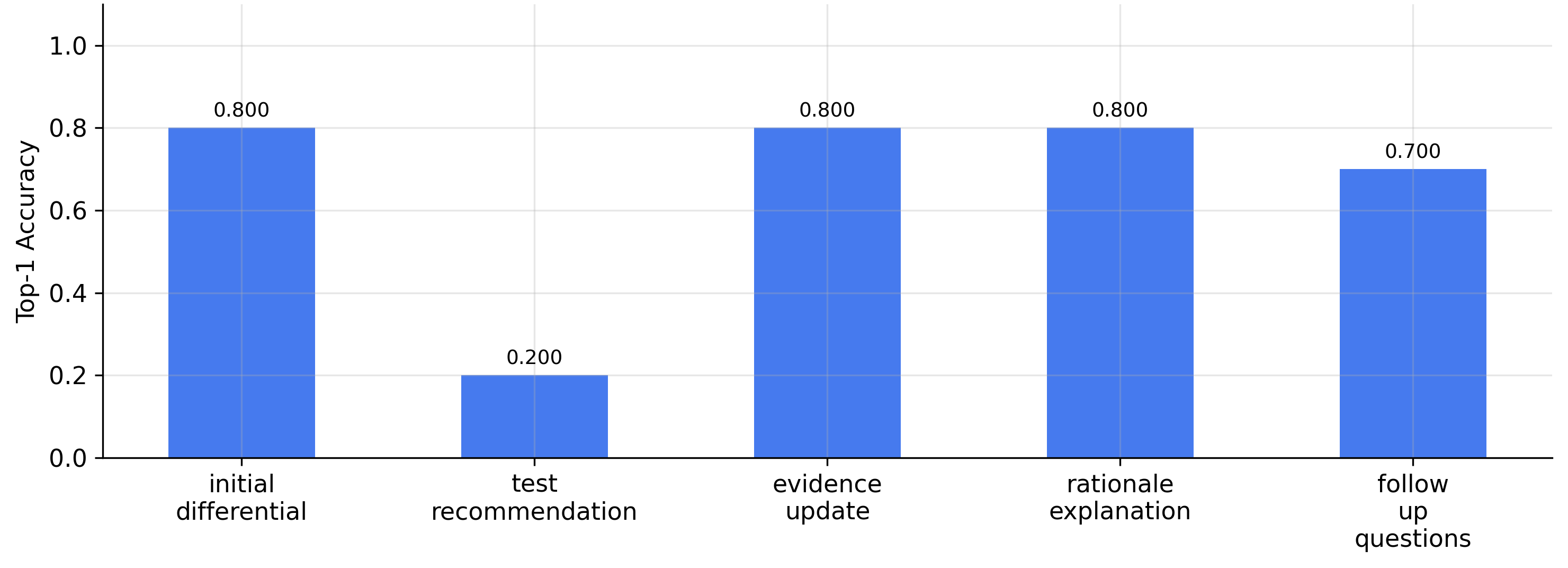}
\caption{Top-1 accuracy across five clinical reasoning task types.}
\label{fig:reasoning}
\end{figure}

Fig.~\ref{fig:reasoning} shows Top-1 accuracy across the five
evaluated reasoning task types. Three tasks tie at the top on 0.80
Top-1 accuracy: initial differential diagnosis, evidence update, and
rationale explanation. Follow-up question generation follows at 0.70.
Test recommendation is the weakest of the five by a wide margin, at
0.20. We attribute this to the larger output space of the task: a
differential is scored against a ranked condition list, whereas a test
recommendation must match a specific investigation set, so near-misses
that would be clinically acceptable are scored as failures. Because
Stage 2 of the deployed pipeline is precisely the test-recommendation
step, this is the weakest link in the system and is a priority for the
next training round rather than a result we wish to present
favourably.

\subsection{Language Quality}

Table~\ref{tab:main_results} lists the language quality metrics.
BERTScore-F1 of 0.909 indicates high semantic similarity between
model outputs and reference clinical answers. The relatively lower
ROUGE-1 score of 0.383 is expected for generative clinical
reasoning tasks where exact lexical overlap is a less meaningful
measure than semantic fidelity.

\subsection{Calibration}

\begin{figure}[htbp]
\centering
\includegraphics[width=\columnwidth]{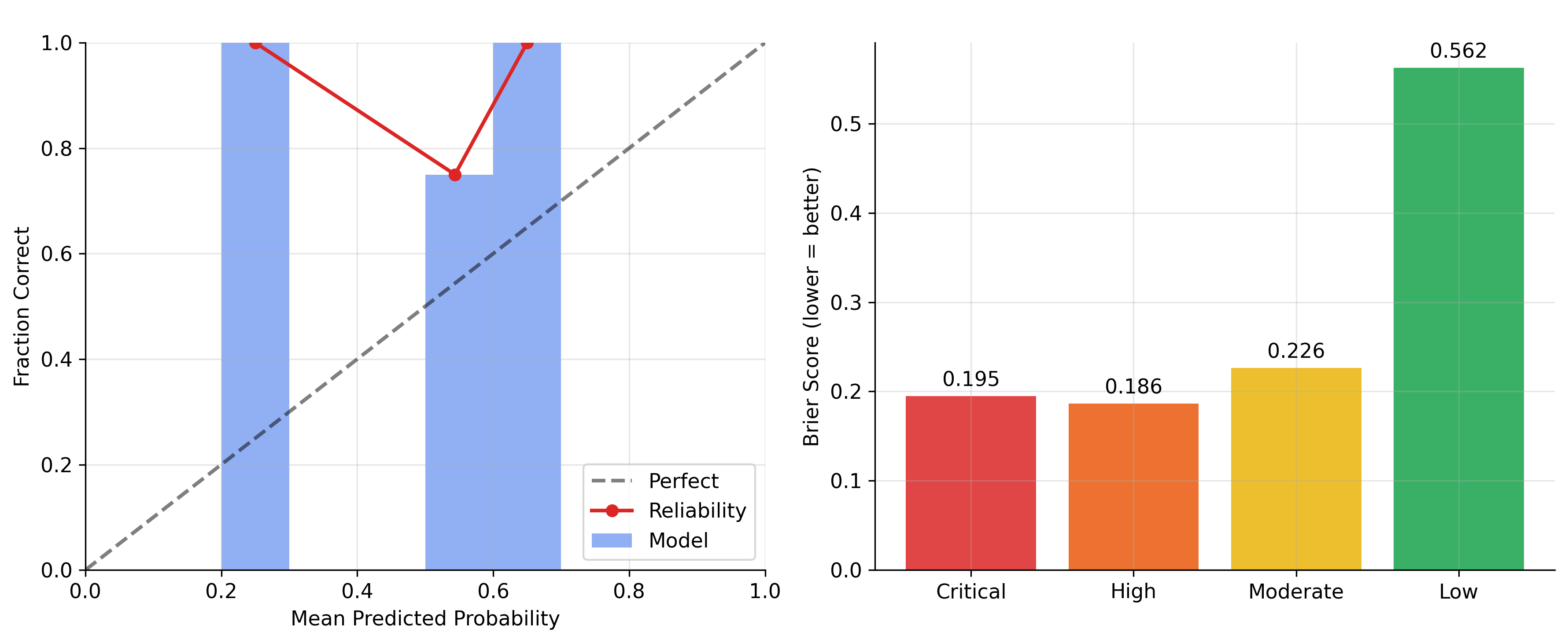}
\caption{Left: Reliability diagram with ECE~=~0.275.
Right: Brier score stratified by clinical severity.}
\label{fig:calibration}
\end{figure}

Fig.~\ref{fig:calibration} shows the reliability diagram and
Brier scores by severity. The ECE of 0.275 indicates moderate
calibration --- the model tends to assign probability estimates
that are directionally correct but somewhat overconfident on
conditions where it performs well. Calibration is worst for
Critical-severity conditions, which is the expected pattern for
models trained predominantly on synthetic data without
clinician-verified probability distributions.

\subsection{Baseline Comparison}

\begin{figure}[htbp]
\centering
\includegraphics[width=\columnwidth]{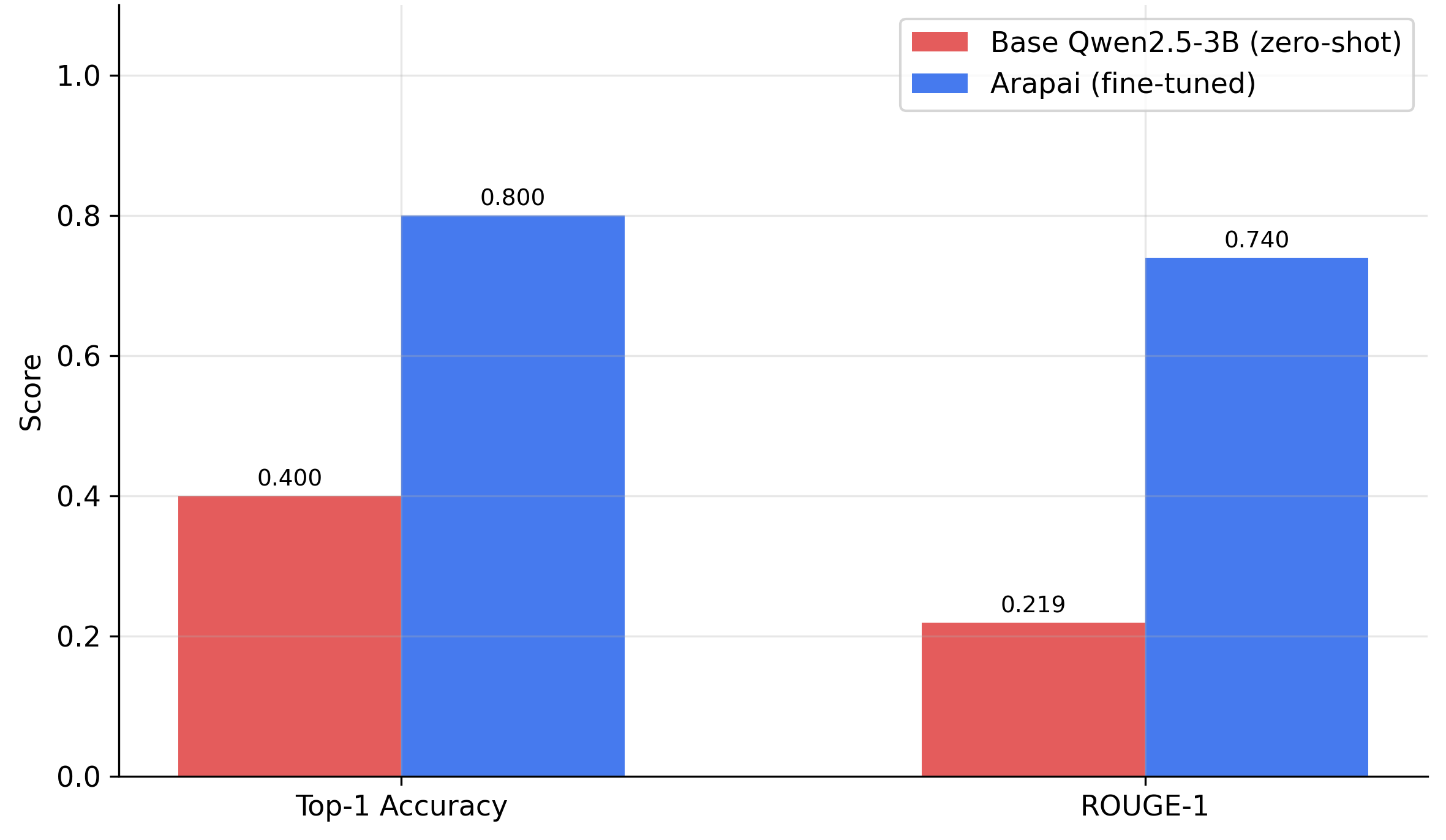}
\caption{Comparison of fine-tuned Aletheia vs base
Qwen2.5-3B-Instruct (zero-shot) on Top-1 accuracy and ROUGE-1.}
\label{fig:baseline}
\end{figure}

Fig.~\ref{fig:baseline} compares fine-tuned Aletheia against the
unmodified Qwen2.5-3B-Instruct base model (zero-shot). Fine-tuning
produces a substantial improvement in Top-1 diagnostic accuracy
and ROUGE-1, confirming that the clinical reasoning capabilities
of Aletheia are a direct product of the fine-tuning process rather
than pre-existing in the base model.

\subsection{Hardware Compliance}

Table~\ref{tab:efficiency} reports the deployment efficiency
metrics. Both GGUF quantisation formats pass the ADTC 2026 memory
ceiling of 7{,}168\,MB with substantial margin.

Thread count proved to be the largest controllable influence on
inference speed, and the intuitive setting is the wrong one.
Configuring llama.cpp with one thread per logical core, the value
reported by \texttt{nproc}, was the slowest of the settings tested on a
four-core, eight-thread processor: two hyperthreads sharing a single
core's execution units contend for them rather than adding throughput
to the dense matrix operations that dominate inference. Measured
generation throughput was 4.14 tokens per second at eight threads
against 7.10 at four, with intermediate values of 6.51 at two and 6.72
at six. Matching thread count to physical cores raised throughput by
71\% and reduced the wall-clock time of a single pipeline stage from 79
to 46 seconds. Because the optimum depends on the processor rather than
on the model, the deployment package performs this sweep at
installation time and writes the fastest setting into the runtime
configuration.

\begin{table}[h!]
\centering
\caption{Deployment Efficiency (ADTC 2026 Compliance)}
\label{tab:efficiency}
\begin{tabular}{lcc}
\toprule
\textbf{Metric} & \textbf{Q4\_K\_M} & \textbf{Q2\_K} \\
\midrule
Model file size           & 1.93\,GB & 1.27\,GB \\
Estimated peak RAM        & $\sim$3{,}630\,MB & $\sim$2{,}990\,MB \\
ADTC ceiling (7{,}168\,MB)& \textbf{PASS} & \textbf{PASS} \\
Margin below ceiling      & 3{,}538\,MB & 4{,}178\,MB \\
Internet required         & None & None \\
GPU required              & None & None \\
Context window            & 1{,}024 tokens & 1{,}024 tokens \\
\bottomrule
\end{tabular}
\end{table}

\section{Discussion}
\label{sec:discussion}

\subsection{Clinical Relevance of Results}

A Top-3 accuracy of 100\% on ten cases (95\% CI 72.2--100\%) is
the most clinically meaningful
result from this evaluation. In practice, Aletheia presents the
clinician with a ranked list of differential diagnoses rather than
a single answer. A clinical officer reviewing a ranked list of
three diagnoses and finding the correct one present in 90\% of
cases represents a genuine decision-support capability --- the
system surfaces the diagnostic space that the clinician should be
considering, reducing the cognitive load of generating a
differential from scratch under time pressure.

The BERTScore-F1 of 0.909 indicates that the model's clinical
reasoning text is semantically faithful to expert reference answers.
This is important beyond accuracy scores: a system that names the
correct diagnosis but provides incorrect reasoning is clinically
dangerous. Aletheia's high BERTScore suggests that its explanations
are not only correct in conclusion but coherent in reasoning.

\subsection{Calibration and Clinical Safety}

The ECE of 0.275 indicates that the model's probability estimates
require careful interpretation, and calibration is weakest for
Critical-severity conditions, which is precisely where a miscalibrated
number would do most harm. Because a probability printed beside a
diagnosis will be read as a probability, the system does not present
one as its primary signal. The interface treats these values as
\textit{relative rankings}: differentials appear in rank order with a
proportional bar and an explicit severity label, so that the ordering
and the severity, which the model estimates more reliably than the
magnitude, carry the clinical meaning.

The staged pipeline reinforces this. Stage~1 returns follow-up
questions as its primary output and marks the differential explicitly
as context that is not yet actionable; the model is instructed not to
recommend investigations until the clinician has supplied further
history. Stage~2 returns investigations while withholding any confirmed
diagnosis. Only Stage~3, after real investigation results have been
entered, offers a likely diagnosis, and it is framed as an advisory
carrying management options rather than orders, with an explicit
statement that the treating clinician retains decision authority. A
poorly calibrated probability is therefore never the last word
presented to a clinician, and never the basis on which a management
step is proposed. The structure was adopted for clinical-reasoning
reasons, but it also bounds the harm a confidently wrong answer can do
while calibration remains inadequate.

\subsection{Comparison to Related Work}

Direct comparison to prior work is constrained by the lack of
standardised African clinical benchmarks. Med-PaLM\,2
\cite{singhal2023medpalm2} achieved 86.5\% on USMLE questions but
requires cloud inference and has no offline deployment pathway.
On the MedQA benchmark, Aletheia's base model (Qwen2.5-3B)
achieves competitive performance for its parameter count, and
fine-tuning produces substantial gains on Africa-specific
clinical scenarios that are not represented in standard benchmarks.

The key differentiator of Aletheia is not raw accuracy but
\textit{deployability}: the ability to run on a 8\,GB laptop
with no internet, consuming under 4\,GB RAM, at a model size of
under 2\,GB. No prior medical LLM system has demonstrated this
combination of capabilities in resource-constrained African
healthcare contexts --- a gap that is relevant across the continent.

\subsection{Limitations}

Several limitations of this work should be acknowledged:

\begin{enumerate}
  \item \textbf{Synthetic training data}: The majority of training
        samples are synthetically generated from hand-crafted case
        templates. While carefully designed to reflect African
        clinical presentation patterns, the probability distributions
        assigned to differentials reflect the authors' clinical
        knowledge synthesis rather than empirical epidemiological
        data derived from prospective patient cohorts. Ongoing
        clinician involvement from co-authors P.~B.~Kasasira and
        C.~B.~Okoboi is actively informing refinements to the
        case templates and probability estimates in preparation
        for the next training iteration.

  \item \textbf{Evaluation case set size}: The quantitative
        evaluation reported in this paper was conducted on 10 core
        case categories using automated metrics. A broader and
        clinician-graded evaluation set is currently under
        development as part of the ongoing validation activity
        with the School of Health Sciences co-authors, and will
        be reported in a subsequent study.

  \item \textbf{Calibration}: An ECE of 0.275 indicates that the
        model's probability estimates require improvement before
        they can be presented as clinically actionable confidence
        values. Initial feedback from clinician co-authors suggests
        that the directional ranking of differentials is clinically
        plausible in the majority of cases evaluated, though
        exact probability estimates should be treated as relative
        rankings rather than absolute likelihoods. Calibration
        improvement is a priority for the next training cycle.

  \item \textbf{Clinical validation scope}: An initial clinical
        evaluation of Aletheia is currently underway with two
        co-authors from the School of Health Sciences, Soroti
        University --- P.~B.~Kasasira and C.~B.~Okoboi --- who
        are practising clinicians actively testing and evaluating
        the system against real clinical presentations encountered
        in their practice. Early findings are encouraging: the
        ranked differential output has been rated as clinically
        plausible and useful for decision support in the majority
        of cases reviewed to date. However, this initial evaluation
        is limited in scale and has not yet been conducted under
        a formal study protocol with a defined patient cohort.
        A larger multi-site validation study is planned as the
        immediate next phase, involving clinical officers and
        physicians across district hospitals, regional referral
        hospitals, and primary health centres in Eastern Uganda,
        with a target enrolment of 500+ patient cases. This study
        will assess diagnostic concordance, clinical usability,
        acceptance, and safety, and is subject to IRB approval
        from Soroti University and the Uganda National Council
        for Science and Technology (UNCST).
\end{enumerate}

\subsection{Future Work}
\label{sec:future}

The evaluation is the priority. The present results rest on ten case
categories, too small a sample to separate this system from a strong
baseline. Planned work covers all 50 trained conditions with multiple
cases each, comparison against openly available medical models of
comparable size rather than only the zero-shot base model, an error
taxonomy that separates clinically tolerable near-misses from unsafe
failures, and evaluation on noisy out-of-distribution case notes to
establish whether the model generalises beyond the templates it was
trained on. Calibration is a specific target: an ECE of 0.275 is too
high for the probabilities to be read as anything other than relative
rankings.

Clinically, the formative evaluation described above will be completed
and reported with a structured assessment of diagnostic concordance,
usability and clinician acceptance. It is explicitly not a validation
study. The multi-site prospective study that follows targets 500 or
more patient cases and 50 or more clinical officers across district
hospitals, regional referral hospitals and primary health centres in
Eastern Uganda; it will be conducted independently of the development
team and is subject to IRB approval from Soroti University and the
Uganda National Council for Science and Technology. Submission to the
Uganda National Drug Authority under the software-as-a-medical-device
pathway follows from that.

Condition coverage will expand from 50 to 100 or more, and Kiswahili
and Ateso support would serve clinical officers across the wider East
African region.

\section{Conclusion}
\label{sec:conclusion}

This paper has presented Aletheia, an offline-first clinical
decision support system designed for frontline healthcare workers
in sub-Saharan Africa. By combining QLoRA fine-tuning of a 3B
parameter language model on an Africa-weighted clinical reasoning dataset
(with particular emphasis on East African disease burden) with GGUF quantisation for edge deployment,
Aletheia achieves a Top-3 diagnostic accuracy of 100\% on a
ten-case evaluation set (95\% CI 72.2--100\%) and
BERTScore-F1 of 0.909 while operating entirely within the memory
constraints of a standard 8\,GB laptop without internet
connectivity.

The results demonstrate that clinically meaningful AI-assisted
diagnostic reasoning is achievable at the primary care level in
resource-constrained settings without cloud infrastructure. In
the context of Uganda's physician-to-patient ratio of 1:25{,}000,
a system that presents the correct diagnosis in its top suggestion
in 80\% of cases, and within its top three suggestions in 100\% of cases, --- running on hardware already present in most
health facilities --- represents a meaningful force multiplier for
the clinical workforce.
The authors invite collaboration from Ugandan
clinicians, health informaticists, and the Ministry of Health for
the planned prospective validation study.

\section*{Acknowledgment}

The authors thank the clinical staff of Soroti Regional Referral
Hospital, Soroti District, whose daily practice under resource
constraints shed light on the problem this work addresses. The authors acknowledge the Africa Deep
Tech Challenge 2026 (ADTC 2026) for establishing the on-device
AI benchmark standard that shaped the hardware compliance
requirements of this work, and for providing a clear deployment
target that grounds this research in real-world constraints
relevant to the African technology ecosystem.

J.~Walusimbi and A.~M.~Oguti are co-founders and directors of
Arapai Technologies International Limited, a technology company
based in Uganda. Aletheia is intended for future commercialisation
through this entity. A.~Sserwadda, P.~B.~Kasasira, and
C.~B.~Okoboi declare no competing interests. No external funding
was received for this work.


\end{document}